\documentclass[eat,twocolumn]{jmlr}

\usepackage{longtable}

\usepackage{booktabs}
\usepackage[load-configurations=version-1]{siunitx} 

\theorembodyfont{\upshape}
\theoremheaderfont{\scshape}
\theorempostheader{:}
\theoremsep{\newline}

\jmlrvolume{}
\firstpageno{1}

\jmlryear{2022}
\jmlrworkshop{Machine Learning for Health (ML4H) 2022}

\DeclareMathOperator*{\argmax}{max}
\DeclareMathOperator*{\argmin}{min}

\usepackage[normalem]{ulem}
\useunder{\uline}{\ul}{}
\usepackage{adjustbox}
\usepackage{multirow}
\usepackage{caption} 
\title[Adversarial Masking for ECG Data]{Pretraining ECG Data with Adversarial Masking Improves Model Generalizability for Data-Scarce Tasks}

  \author{\Name{{Jessica Y.} Bo} \Email{jbo@mit.edu} \\
  \Name{{Hen-Wei} Huang} \Email{henwei@mit.edu}\\
  \addr Dept. of Medicine, Harvard Medical School, Boston, United States \\ \\
  \Name{Alvin Chan} \Email{alvincgw@mit.edu} \\ 
  \Name{Giovanni Traverso} \Email{cgt20@mit.edu}\\
  \addr Dept. of Mechanical Engineering, Massachusetts Institute of Technology, Cambridge, United States}

\begin{document}

\maketitle

\begin{abstract}
Medical datasets often face the problem of data scarcity, as ground truth labels must be generated by medical professionals. One mitigation strategy is to pretrain deep learning models on large, unlabelled datasets with self-supervised learning (SSL). Data augmentations are essential for improving the generalizability of SSL-trained models, but they are typically handcrafted and tuned manually. We use an adversarial model to generate masks as augmentations for 12-lead electrocardiogram (ECG) data, where masks learn to occlude diagnostically-relevant regions of the ECGs. 
Compared to random augmentations, adversarial masking reaches better accuracy when transferring to to two diverse downstream objectives: arrhythmia classification and gender classification. Compared to a state-of-art ECG augmentation method \textit{3KG}, adversarial masking performs better in data-scarce regimes, demonstrating the generalizability of our model\footnote{Our code is available at: \url{https://github.com/jessica-bo/advmask_ecg}}.
\end{abstract}

\section{Introduction}
\label{sec:intro}
Across medical applications, deep learning is increasingly used to automate disease diagnosis \citep{Miotto2018DeepChallenges}.
In some cases, neural networks have even reached or exceeded the performance of expert physicians  \citep{Hannun2019Cardiologist-levelNetwork}. 
One such application is with 12-lead electrocardiogram (ECG) data, which is commonly collected to screen for various cardiovascular disorders \citep{Fesmire1998UsefulnessPain}. There has been a surge of deep learning ECG research enabled by large scale challenges like PhysioNet/Computing in Cardiology \citep{Reyna2021Will2021}. 
While these results show substantial progress of the field, their performance only reflects training and testing across large-scale, labelled dataset. In contrast, real world medical datasets are likely small due to the extensive resources required to collect medical labels.




Data scarcity is a well-documented issue that effect deep learning training. Models trained on small datasets lack generalizability to unseen data and cannot be deployed reliably \citep{Kelly2019KeyIntelligence}. 
To circumvent small datasets, large yet unlabelled dataset can be leveraged to pretrain deep learning models with robust representations, which is commonly done in the ECG domain \citep{Sarkar2020, Weimann2021, Liu2021Self-supervisedPre-training, Kiyasseh2021CLOCS:Patients, Diamant2022, Mehari2022, Oh2022Lead-agnosticElectrocardiogram}. 
A pretraining technique that does not require explicit labels is self-supervised learning (SSL), such SimCLR \citep{Chen2020ARepresentationsb}, where the encoding model is trained to maximize the similarity between augmented pairs of data.

For time-series data, augmentations do not reach uniform and predictable gains across different datasets and tasks, making the augmentation selection process difficult \citep{Iwana2021}. Augmentations are typically selected from a standard pool like noise injection, baseline shift, time-domain masking, and more recent advances using generative adversarial models (GANs) to create synthetic data \citep{Wen2021}. 
A recent work \textit{3KG} develops physiologically-consistent spatial augmentations for ECGs that reaches good performance on small datasets \citep{Gopal20213KG:Augmentations}. 
However, it is handcrafted with manually tuned parameters, while we investigate the use of an adversarial method to optimize the augmentation. 

In this work, we elect to focus on time-domain masking, as it is underexploited for periodic time-series data like ECGs: 
\begin{itemize}
    \item We adapt the image-based adversarial masking method of \citet{Shi2022AdversarialLearning} to the time domain and generate masks as ECG augmentations for SSL pretraining. To our knowledge, this is the first work to implement adversarial masking for time-series data. 
    \item We show that incorporating adversarial masking in SSL pretraining improves performance in a downstream classification tasks compared to random augmentations. Adversarial masking also improves downstream performance in data-scarce regimes compared to the state-of-art augmentation \textit{3KG} and a statistical baseline \textit{Peakmask}.
\end{itemize}




\section{Methods}
\label{sec:methods}

The adversarial masking pretraining framework with the encoding model $\boldsymbol{E}$ and adversarial masking model $\boldsymbol{A}$ and the downstream transfer learning step is show in Figure \ref{fig:advaug_model}.

\subsection{Self-Supervised Pretraining}
\label{sec:ssl}

\paragraph{SimCLR Objective:} SimCLR is an SSL framework that learns to align representations of pairs of augmented data in a contrastive manner. Given a batch of data $\left\{\boldsymbol{x_{i}}\in \mathcal{R}^D\right\}_{i=1}^{B}$, an augmented view of $\boldsymbol{x}_{i}$ can be created through a collation of one or more augmentations $T(\boldsymbol{x}_{i})$. The encoding model $\boldsymbol{E}$ is trained to align the feature representations of positive pairs $\boldsymbol{h}_{i}=\boldsymbol{E}(\boldsymbol{x}_{i})$ and $\boldsymbol{h'}_{i}=\boldsymbol{E}(T(\boldsymbol{x}_{i}))$, while separating them from all other negative pairs within the batch, where $\boldsymbol{x}_{i} \neq \boldsymbol{x}_{j}$. The simplified SimCLR objective is described in Equation \ref{eq:nce}\footnote{The temperature parameter $\tau$ is omitted for simplicity but takes the value of 0.1.}.

\begin{equation}
\label{eq:nce}
    \mathcal{L}_{\text {SSL}}(\boldsymbol{x} ; \boldsymbol{E}) = -
    \log \frac{\exp (sim(\boldsymbol{h}_{i}, \boldsymbol{h'}_{i})) }{\sum_{i \neq j} \exp (sim \left ( \boldsymbol{h}_{i}, \boldsymbol{h'}_{j} \right ) )}
\end{equation}

\paragraph{Adversarial Objective:} We add an adversarial masking model $\boldsymbol{A}$ that generates a set of \textit{N} masks for a given data sample, $\boldsymbol{m}_{i} \in \mathcal{R}^{N \times D}=\boldsymbol{A}(\boldsymbol{x}_{i})$. Multiple masks ensure that the masked regions of the ECG are alternated, as only one mask is sampled, binarized, and applied in each training step. 
We use a differentiable soft binarization function to drive the mask values towards either 0 or 1, shown in Equation \ref{eq:binarize} with $\gamma=25$. This enables back-propagation while simulating a shape-preserving binary mask without significant distortions to the ECG signals.

\begin{equation}
\label{eq:binarize}
    \boldsymbol{m}_{binarized} = \frac{1}{1 + \exp( -\gamma (\boldsymbol{m} - 0.5))}
\end{equation}

The masking model $\boldsymbol{A}$ acts in opposition to the encoding model $\boldsymbol{E}$ by generating difficult augmentations, hence maximizing the SimCLR objective. We also adapt the sparse penalty from \citet{Shi2022AdversarialLearning} to limit the amount of masking, as described by Equation \ref{eq:sparse} with $\alpha=0.1$ as a weighting term. The full \textit{min-max} loss is described by Equation \ref{eq:minmax}: 

\begin{equation}
\label{eq:sparse}
    \mathcal{L}_{\text {sparse}}(\boldsymbol{x} ; \mathcal{A})=\sin \left(\frac{\pi}{D} \sum_{d=1}^{D} \boldsymbol{m}^{d}\right)^{-1}
\end{equation}

\begin{equation}
\label{eq:minmax}
\argmin\limits_{\boldsymbol{E}} \argmax\limits_{\boldsymbol{A}} 
     \mathcal{L}_{\text {SSL}}(\boldsymbol{E}, \boldsymbol{A}) -
    \mathcal{L}_{\text {sparse}}(\boldsymbol{A})
\end{equation}

\begin{figure}[htbp]
\floatconts
  {fig:advaug_model}
  {\caption{Adversarial masking pretraining and downstream transfer phases.}}
  {\includegraphics[width=0.95\linewidth]{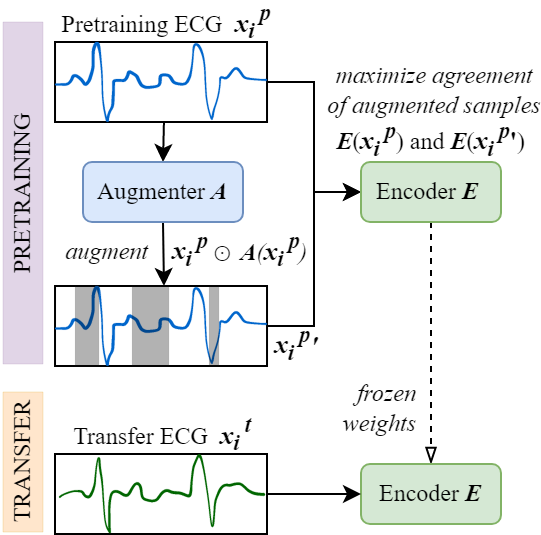}}
\end{figure}

\paragraph{Architectures:} For the encoder backbone, we use a 1D ResNet-18 with a hidden dimension of 512 \citep{He2016DeepRecognition}, which is the best performing architecture in multiple ECG tasks \citep{Nonaka2021In-depthDiagnosis}. We use a two-layer projection head typical in contrastive SSL to convert the encoder's outputs to a 128-dimension space, but the projector is not transferred to the downstream task.

For the adversarial model, we adapt an open-source 1D U-Net with four downsampling and upsampling layers. 
The number of out channels is the number of masks \textit{N}. The outputs are passed through either a \textit{softmax} or \textit{sigmoid} function depending on \textit{N}.

\paragraph{Pretraining Details:} All models are implemented with PyTorch Lightning. Training and testing are performed with a NVIDIA Volta V100 GPU on the MIT Supercloud \citep{reuther2018interactive}. We pretrain on the 12-lead ECG datasets from the PhysioNet/Computing in Cardiology Challenge 2021 \citep{Reyna2021Will2021}, which comprises over 88 thousand patient recordings from seven institutions in four countries. 

We use a learning rate of 0.0001 and an \textit{Adam} optimizer for both the encoding model and adversarial model. The models are updated one after another with their respective losses within the same batch. The batch size is 32 with gradient accumulation over 4 batches, which is equal to an effective batch size of 128. To reduce computation efforts, we use mixed precision training. An early stopping condition based on minimizing $\mathcal{L}_{\text {SSL}}$ is implemented to reduce the training time.


\subsection{Downstream Transfer Learning}
\label{sec:transfer}
We transfer the encoding models's learned representations to two downstream classification tasks by training a linear layer on top of the frozen model weights. The tasks, arrhythmia classification and gender classification, are used as benchmarks by \citet{Diamant2022}. The former correlates to disease characteristics and the latter to patient identity, representing two diverse objectives. 

The 12-lead Chapman-Shaoxing ECG dataset comprising over 10 thousand patients and four classes of cardiac rhythm labels is used for transfer \citep{Zheng2020APatients}. The arrhythymia class balance is 36\% atrial fibrillation (AFIB), 22\% supraventricular tachycardia (GSVT), 21\% sinus bradycardia (SB), and 21\% sinus rhythm (SR). The gender balance is 56\% male and 44\% female. 

Transfer learning is performed with a batch size of 256, learning rate of 0.01, and an \textit{Adam} optimizer. A linear layer projects the output dimension of the encoding model to the number of classes.

\section{Results}
\label{sec:results}

\subsection{Transfer Task Performance}
\label{sec:performance}

{\renewcommand{\arraystretch}{1.05}
\begin{table*}
\centering
  {\caption{Downstream performance for two tasks across multiple training dataset sizes (\textit{X\% DS}) reported for encoders pretrained on multiple augmentations and a \textit{Scratch} baseline. }
  \label{tab:results}}
  \resizebox{\linewidth}{!}{
    \begin{tabular}{ c c c c c c c }
     \toprule
       & \multicolumn{3}{c}{{Arrhythmia Classification (\%)}}& \multicolumn{3}{c}{{Gender Classification (\%)}}  \\
      & \textit{100\% DS}& \textit{10\% DS}& \textit{1\% DS}& \textit{100\% DS}& \textit{10\% DS}& \textit{1\% DS}\\
     \hline
     {\textit{Scratch}} & \textit{70.99 \tiny $\pm$ 1.59} & \textit{63.92 \tiny $\pm$ 1.97} & \textit{34.89 \tiny $\pm$ 0.54} &\textit{ 68.18 \tiny $\pm$ 1.52 }&\textit{ 56.26 \tiny $\pm$ 4.24} & \textit{50.78 \tiny $\pm$ 4.54}\\
     \hline
     {Gaussian} & 68.96 \tiny $\pm$ 2.05 & 65.52 \tiny $\pm$ 3.47 & 49.69 \tiny $\pm$ 2.01 & 65.68 \tiny $\pm$ 1.34 & 64.05 \tiny $\pm$ 0.753 & 60.39 \tiny $\pm$ 1.38\\
     {Powerline} & 72.93 \tiny $\pm$ 1.93 & 69.27 \tiny $\pm$ 1.11 & 55.29 \tiny $\pm$ 2.05 & 68.59 \tiny $\pm$ 1.49 & 66.83 \tiny $\pm$ 2.5 & 60.73 \tiny $\pm$ 1.83\\
     {STFT} & 73.0 \tiny $\pm$ 0.978 & 70.21 \tiny $\pm$ 1.18 & 53.75 \tiny $\pm$ 0.96 & 68.4 \tiny $\pm$ 0.742 & 67.49 \tiny $\pm$ 0.734 & 62.01 \tiny $\pm$ 3.16\\
     {Wander} & 75.53 \tiny $\pm$ 2.23 & 73.06 \tiny $\pm$ 1.98 & 58.38 \tiny $\pm$ 1.91 & 68.55 \tiny $\pm$ 0.414 & 65.83 \tiny $\pm$ 1.84 & 62.27 \tiny $\pm$ 2.46\\
     {Shift} & 76.38 \tiny $\pm$ 1.64 & 74.75 \tiny $\pm$ 1.39 & 58.76 \tiny $\pm$ 3.27 & 69.27 \tiny $\pm$ 0.854 & 68.74 \tiny $\pm$ 0.392 & 61.64 \tiny $\pm$ 3.45\\
     {Mask}& 72.97 \tiny $\pm$ 5.19 & 69.05 \tiny $\pm$ 4.84 & 54.85 \tiny $\pm$ 4.86 & 69.49 \tiny $\pm$ 1.37 & 68.02 \tiny $\pm$ 2.71 & 64.64 \tiny $\pm$ 1.2\\
     {Blockmask} & 81.76 \tiny $\pm$ 0.454 & 78.32 \tiny $\pm$ 1.45 & 65.3 \tiny $\pm$ 3.39  & 68.3 \tiny $\pm$ 2.56 & 66.8 \tiny $\pm$ 0.769 & 63.05 \tiny $\pm$ 1.43\\
     {Peakmask} & 85.67 \tiny $\pm$ 0.64 & 79.29 \tiny $\pm$ 1.72 & 62.33 \tiny $\pm$ 3.87 & \textbf{74.44 \tiny $\pm$ 0.66 }& \textbf{72.53 \tiny $\pm$ 1.21} & 61.58 \tiny $\pm$ 7.14\\
     {\bfseries Adv Mask (AM)} & 85.14 \tiny $\pm$ 0.51 & 81.82 \tiny $\pm$ 0.85& 72.59 \tiny $\pm$ 2.99 &  70.53 \tiny $\pm$ 0.51 & 70.49 \tiny $\pm$ 0.63 & \textbf{67.24 \tiny $\pm$ 1.45}\\
     \hline
     {3KG} & 86.39 \tiny $\pm$ 2.42 & 78.75 \tiny $\pm$ 2.71 & 70.65 \tiny $\pm$ 4.96 & 64.14 \tiny $\pm$ 2.06 & 64.27 \tiny $\pm$ 1.63 & 60.42 \tiny $\pm$ 1.87\\
     {\bfseries AM + 3KG} & \textbf{88.77 \tiny $\pm$ 1.31} &\textbf{ 86.61 \tiny $\pm$ 0.78 }&\textbf{ 79.88 \tiny $\pm$ 1.25 }& 70.49 \tiny $\pm$ 0.72 & 68.46 \tiny $\pm$ 1.83 & 62.3 \tiny $\pm$ 2.57\\
     \bottomrule
    \end{tabular}
    }
\end{table*}

We present transfer learning results with a \textit{train-validation-test }split of 80\%, 10\% and 10\%. We also simulate real life data scarcity conditions by reducing the transfer training dataset to various fractions of the original size (8516 samples). The test set is held constant and the average test accuracy for all trials is reported across 3 random seeds. 

Table \ref{tab:results} shows the prediction accuracy and standard deviation of both classification tasks with models pretrained on all baseline and adversarial augmentations. Baseline augmentations taken from literature and details are available in Appendix \ref{apd:first}. For each task at each dataset size reduction, the best performing augmentation is bolded. Results from a finetuned \textit{Scratch} baseline is also reported. Adversarial masking (\textbf{Adv Mask}) results are given with $\textit{N}=2$. 

Across both tasks, \textbf{Adv Mask} yields superior results to most baseline augmentations. In arrhythymia classification, \textbf{Adv Mask} is competitive to \textit{3KG} \citep{Gopal20213KG:Augmentations}, a complex spatiotemporal augmentation specific to ECGs, and even outperforms it in data scarce regimes. Ultimately, the best performance is achieved by combining \textbf{Adv Mask + 3KG}, showing the orthogonal benefits of our methods. In gender classification, \textbf{Adv Mask} maintains its advantage in low-data trials, but otherwise falls to \textit{Peakmask}, which is introduced by us and not a standard augmentation. 
It is computed statistically by masking out peaks in the signal, which is visually similar to what \textbf{Adv Mask} achieves, see Figure \ref{fig:masks}. 

Figure \ref{fig:DS_size} shows a more fine-grained graph of how training dataset size affects classification accuracy for the top performing augmentations (\textit{Peakmask}, \textit{3KG}, \textbf{Adv Mask}, \textbf{Adv Mask + 3KG}) and \textit{Scratch}.

\begin{figure}[htbp]
\floatconts
  {fig:DS_size}
  {\caption{Downstream classification accuracy at different training dataset sizes.}}
  {\includegraphics[width=0.95\linewidth]{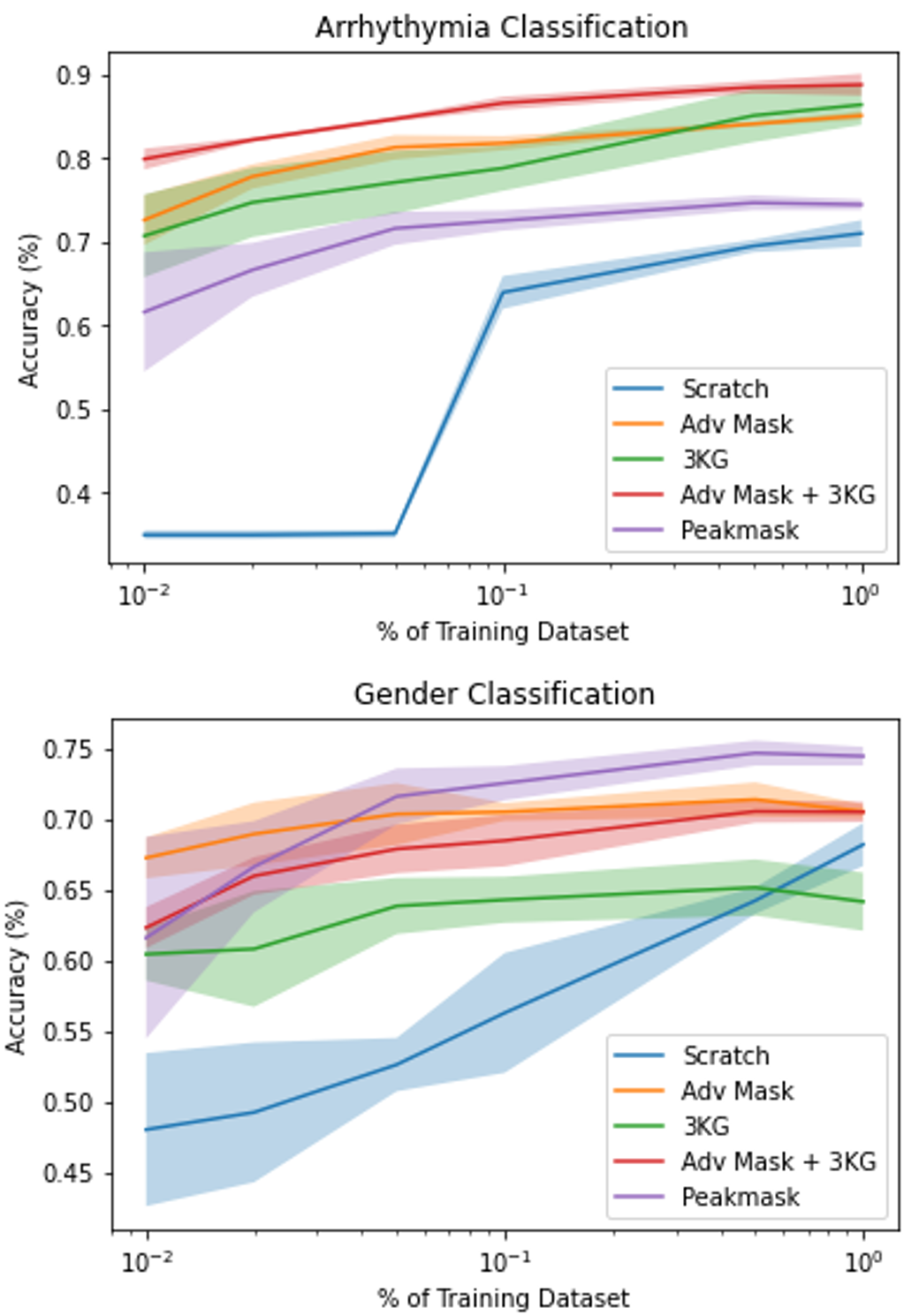}}
\end{figure}

\subsection{Analysis of Augmentations}
\label{sec:augmentations}

Figure \ref{fig:masks} visualizes the generated masks, with more examples in Appendix \ref{apd:fourth}. Lead II of the ECG is displayed in red and the masked signal in green. Since we use a soft binarization method, the mask can take any value between 0 and 1. The blue overlay represent the regions that are heavily driven to dropout. The shape of the original ECG is well preserved outside of the masked regions, but may be augmented with a slight shift and scaling factor. \citet{Huang2021TowardsLearning} show richer data augmentations lead to higher model generalizability in downstream tasks, which supports why \textbf{Adv Mask} is effective. 

\begin{figure}[htbp]
\floatconts
  {fig:masks}
  {\caption{\textit{N}=2 adversarial masks.}}
  {\includegraphics[width=0.9\linewidth]{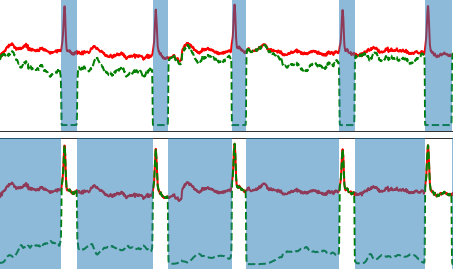}}
\end{figure}

With  $\textit{N}=2$, one mask consistently covers the QRS complex, while the other mask covers the remaining areas. We hypothesize this is similar to capturing diagnostically-relevant ``semantic" content of the ECG. 
In salient regions of ECG data, the QRS complex often carry high levels of disease-diagnostic information \citep{Jones2020ImprovingMaps}. 
This suggests that the encoding model emphasizes learning the structual information of the  salient regions during pretraining. While the results may indicate that learnt masks converge to peaks, we show that \textbf{Adv Mask} outperforms \textit{Peakmask} in the arrhythmia classification task and in low-data regimes in the gender classification task. 

\section{Conclusions and Future Work}
\label{sec:conclusions}
Adversarial masking for ECG data can be utilized positively in SSL pretraining of deep learning models. It induces the learning of generalizable representations for downstream transfer learning in highly data-scarce tasks.
Masking as an augmentation scheme is agnostic to the choice of architecture and training scheme, so the method can be extended to other SSL frameworks and time-series data modalities. 

Future work includes evaluating across more datasets to understand when adversarial masking can bring the most benefit. To strengthen our results, we would also like to extend our experiments to include generative models as an augmentation baseline and tailor the sparsity penalty to leverage specific properties of time-series data.

\bibliography{references}

\appendix

\section{Augmentations}\label{apd:first}

The baseline augmentations are sourced from ECG deep learning research but does not cover an exhaustive list. However, we aim to be representative of typical random ECG augmentation techniques. Note that a limitation of our results is that we only test with isolated augmentations, whereas combining two or more augmentations may intuitively yield better results. We also do not have generative models (GANs) as baseline, but we aim to include them in the future.

\paragraph{Gaussian Noise:} A vector of noise $v(t) \sim \mathcal{N}(0, 0.05)$ is sampled and added to each lead of the ECG. Gaussian noise injection is a very common augmentation used in SSL training for both time-series and image data.
\paragraph{Powerline Noise:} Powerline noise is  $n(t) = \alpha cos(2\pi t k f_{p} + \phi)$, with $\alpha \sim \mathcal{U}(0, 0.5)$, $\phi \sim \mathcal{N}(0, 2\pi)$, and $f_{p} = 50 Hz$  \citep{Mehari2022, Oh2022Lead-agnosticElectrocardiogram}.
\paragraph{Short-time Fourier Transform:} STFT involves computing the Fourier transform of short segments of the time-series signal to generate a spectrogram. A random mask with values sampled from a beta distribution $B(\alpha=5, \beta=2)$ is applied to the spectrogram. Finally, the STFT operation is inversed to recover the time domain signal. 
\paragraph{Baseline Wander:} The ECG signal is perturbed with a very low frequency signal to simulate a drift $n(t) = C \sum_{k=1}^{K} a \cos \left(2 \pi t k \Delta f+\phi\right)$,  with $C \sim \mathcal{N}(1, 0.5^{2})$, $\alpha \sim \mathcal{U}(0, 0.5)$, $\phi \sim \mathcal{N}(0, 2\pi)$, and $\Delta f \sim \mathcal{U}(0.01, 0.2)$ \citep{Mehari2022, Oh2022Lead-agnosticElectrocardiogram}.
\paragraph{Baseline Shift:} A fraction $p = 0.2$ of the baseline of each lead of the ECG signal is shifted positively or negatively by a factor of $\alpha \sim \mathcal{N}(-0.5, 0.5)$ \citep{Mehari2022, Oh2022Lead-agnosticElectrocardiogram}.
\paragraph{Mask:} Two types of random masking are implemented, where Mask refers to any timestep being masked out to 0 with the probability $p=0.2$.
\paragraph{Blockmask:} Blockmask masks out a continuous portion $p=0.2$ of each lead to 0. The main difference is that Blockmask would occlude larger structural regions of the ECG, whereas Mask only occludes local details.
\paragraph{Peakmask:} We introduce Peakmask as a statistical baseline that aims to replicate the peak-finding behaviour of the adversarially generated masks. Timesteps where the average value across leads exceeds the average value over the entire ECG are masked. We either sample the computed mask directly, or the opposite regions (non-peak areas). This emulates the Adv Mask technique with N=2. 
\paragraph{3KG:} Developed by \citet{Gopal20213KG:Augmentations}, 3KG augments the 3D spatial representation of the ECG in vectorcardiogram (VCG) space and reprojects it back into ECG space, mimicking natural variations in cardiac structure and orientation. We take the best parameters reported in the paper, a random rotation $-45^{\circ} \leq \theta \leq 45^{\circ}$ and a scaling factor $1 \leq s \leq 1.5$. 

\section{Adversarial Masks}\label{apd:fourth}

Figures \ref{fig:masksN1}-\ref{fig:masksN12} depict more examples of masks with $\textit{N}=1,3,12$. When $\textit{N} < 12$, one mask is randomly sampled and applied to all leads of the ECG. When $\textit{N} = 12$, each mask is applied separately to each lead. Increasing \textit{N} does not seem to be helpful as masks tend to cover either peaks or flat regions, which can be achieved with \textit{N}=2.

\begin{figure}[htbp]
\floatconts
  {fig:masksN1}
  {\caption{\textit{N}=1 adversarial masks.}}
  {\includegraphics[width=0.9\linewidth]{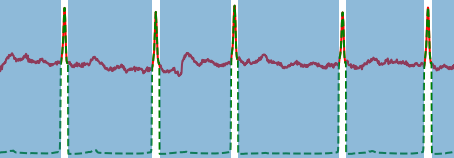}}
\end{figure}

\begin{figure}[htbp]
\floatconts
  {fig:masksN3}
  {\caption{\textit{N}=3 adversarial masks.}}
  {\includegraphics[width=0.9\linewidth]{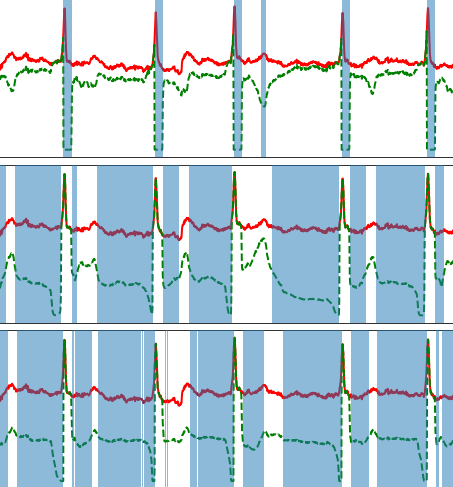}}
\end{figure}

\begin{figure}[htbp]
\floatconts
  {fig:masksN12}
  {\caption{\textit{N}=12 adversarial masks; leads are displayed and masked separately.}}
  {\includegraphics[width=0.9\linewidth]{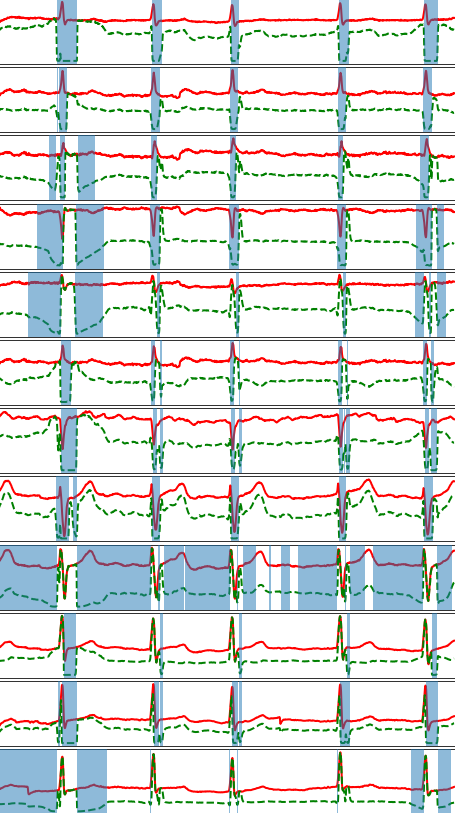}}
\end{figure}

\let\clearpage\relax

\end{document}